\documentclass[
    nolists,
    subscriptcorrection,
    mathalfa=cal=euler,
    balance,
    nofoot
]{asmejour}

\graphicspath{{figures/}}

\JourName{Mechanisms and Robotics}

    \newcommand{\new}[1]{#1}
    
    \newcommand{\cut}[1]{}
    \newcommand{\cutmath}[1]{}
    \newcommand{\cutcitation}[1]{}

\def\norm#1{\left\|{#1}\right\|}
\def\tr{^T}
\DeclareMathOperator{\sign}{sign}
\DeclareMathOperator{\wrapToPi}{wrapToPi}

\usepackage[disable]{todonotes}

\begin{document}

\date{}

\SetAuthorBlock{Alexander J. Elias\CorrespondingAuthor}{
   email: \href{mailto:eliasa3@rpi.edu}{eliasa3@rpi.edu}} 

\SetAuthorBlock{John T. Wen}{%
   email: \href{mailto:wenj@rpi.edu}{wenj@rpi.edu}
}

\SetAuthorBlock{\phantom}{%
Department of Electrical, Computer, and Systems Engineering,\\
   Rensselaer Polytechnic Institute,\\
   Troy, NY 12180, USA
}

\title{Path~Planning~and~Optimization for~Cuspidal~6R~Manipulators}

\begin{abstract}
% RA-L: no more than 200 words
% ASME Journal: No more than 250 words
A cuspidal robot can move from one inverse kinematics (IK) solution to another without crossing a singularity. Multiple industrial robots are cuspidal. They tend to have a beautiful mechanical design, but they pose path planning challenges.
A task-space path may have a valid IK solution for each point along the path, but a continuous joint-space path may depend on the choice of the IK solution or even be infeasible.
This paper presents new analysis, path planning, and optimization methods to enhance the utility of cuspidal robots. We first demonstrate an efficient method to identify cuspidal robots and show, for the first time, that the ABB GoFa and certain robots with three parallel joint axes are cuspidal.
We then propose a new path planning method for cuspidal robots by finding all IK solutions for each point along a task-space path and constructing a graph to connect each vertex corresponding to an IK solution. Graph edges \cut{are weighted}\new{have a weight} based on the optimization metric, such as minimizing joint velocity.
The optimal feasible path is the shortest path in the graph.
This method can find non-singular paths as well as smooth paths which pass through singularities.
Finally, \new{we incorporate }this path planning method \cut{is incorporated }into a path optimization algorithm. Given a fixed workspace toolpath, we optimize the offset of the toolpath in the robot base frame while ensuring continuous joint motion.
Code examples are available in a publicly accessible repository.
\end{abstract}
\maketitle

\section{Introduction}
The growing ubiquity of robotic manipulators has resulted in new kinematic designs with different sets of intersecting or parallel joint axes. This has led to the unexpected problem that classical path planning algorithms fail on certain classes of robots due to a property called cuspidality.

The most common design for large industrial robot arms features a spherical wrist and two parallel axes. Examples include the ABB IRB 6640~\cite{ABB_6640} and the YASKAWA Motoman MA2010-A0~\cite{MA2010}. Cobots designed for collaborative tasks such as the UR5~\cite{UR5} or Techman TM5-700~\cite{techman} feature three parallel axes plus other intersecting axes. In either case, a given end effector pose has up to eight inverse kinematics~(IK) solutions which \cut{are found by solving}\new{correspond to the solutions of} a series of quadratic equations. The IK solutions \cut{can be intuitively labeled}\new{have intuitive labels} and are separated by singularities.

\begin{figure}[t]%
    \centering%
    \begin{subfigure}[t]{0.25\linewidth}%
        {\includegraphics[width=\linewidth]{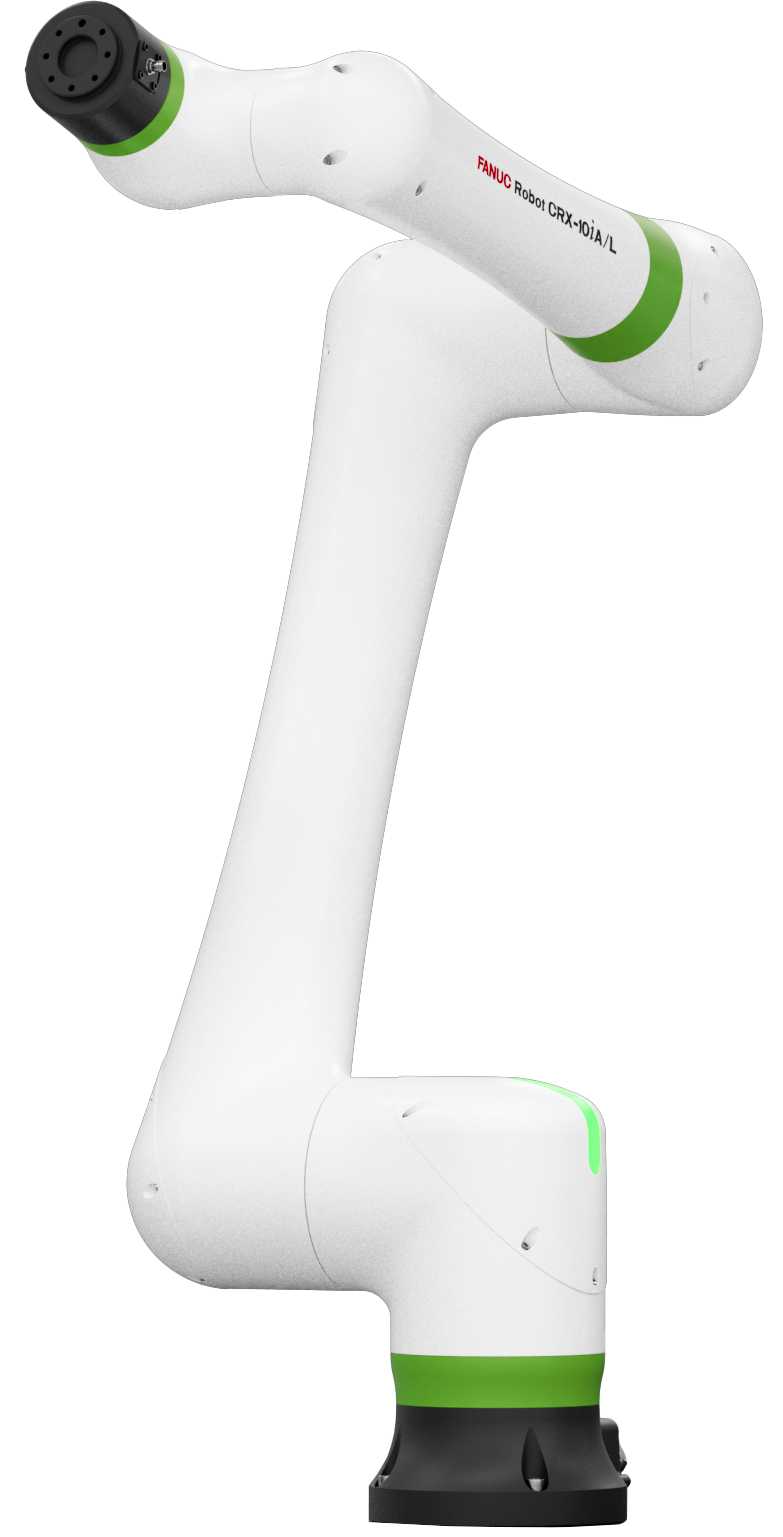}}%
        \subcaption{\label{fig:CRX_photo}}%
    \end{subfigure}%
    \begin{subfigure}[t]{0.35\linewidth}%
        {\includegraphics[width=\linewidth, trim={0in, 0in, 3.8in, 0in},clip]{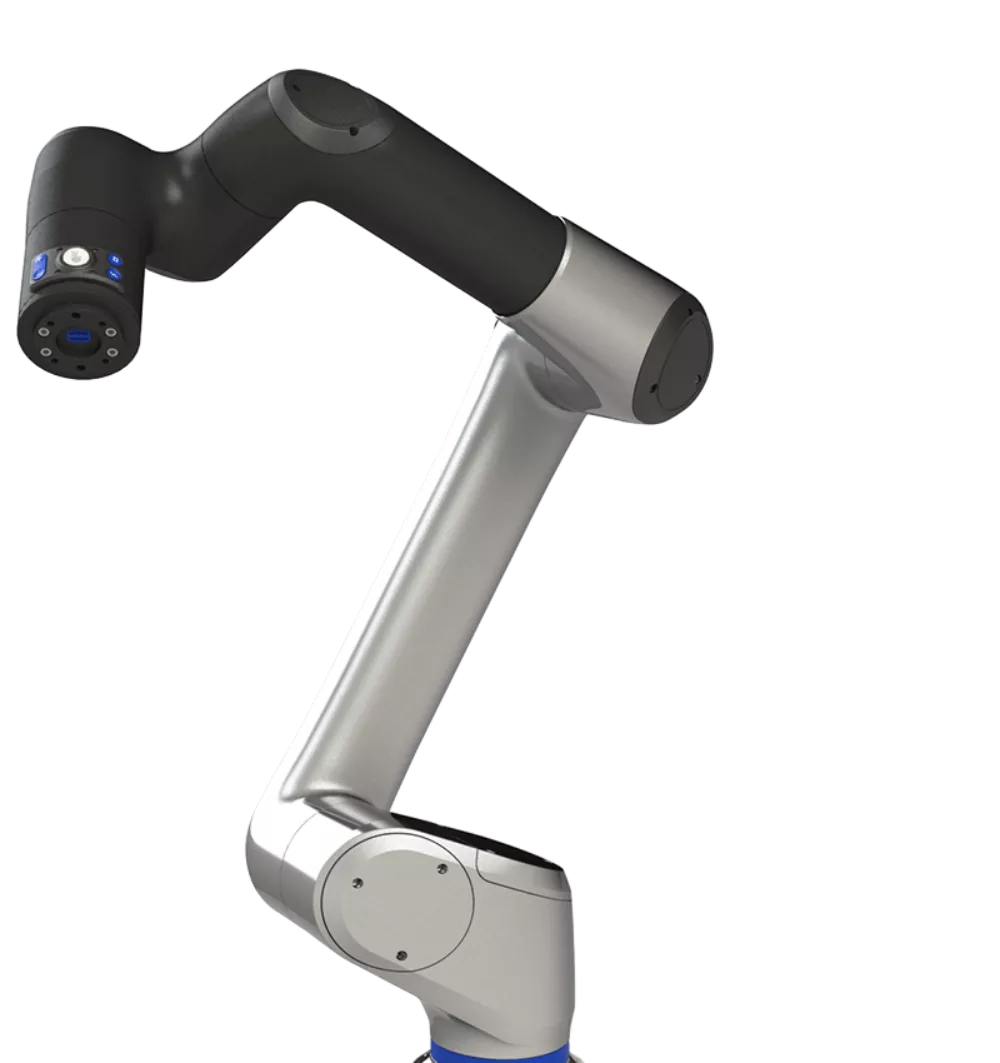}}%
        \subcaption{\label{fig:kinova_photo}}%
    \end{subfigure}%
    \begin{subfigure}[t]{0.4\linewidth}%
        {\includegraphics[width=\linewidth, trim={3in, 2.6in, 5.5in, 4in},clip]{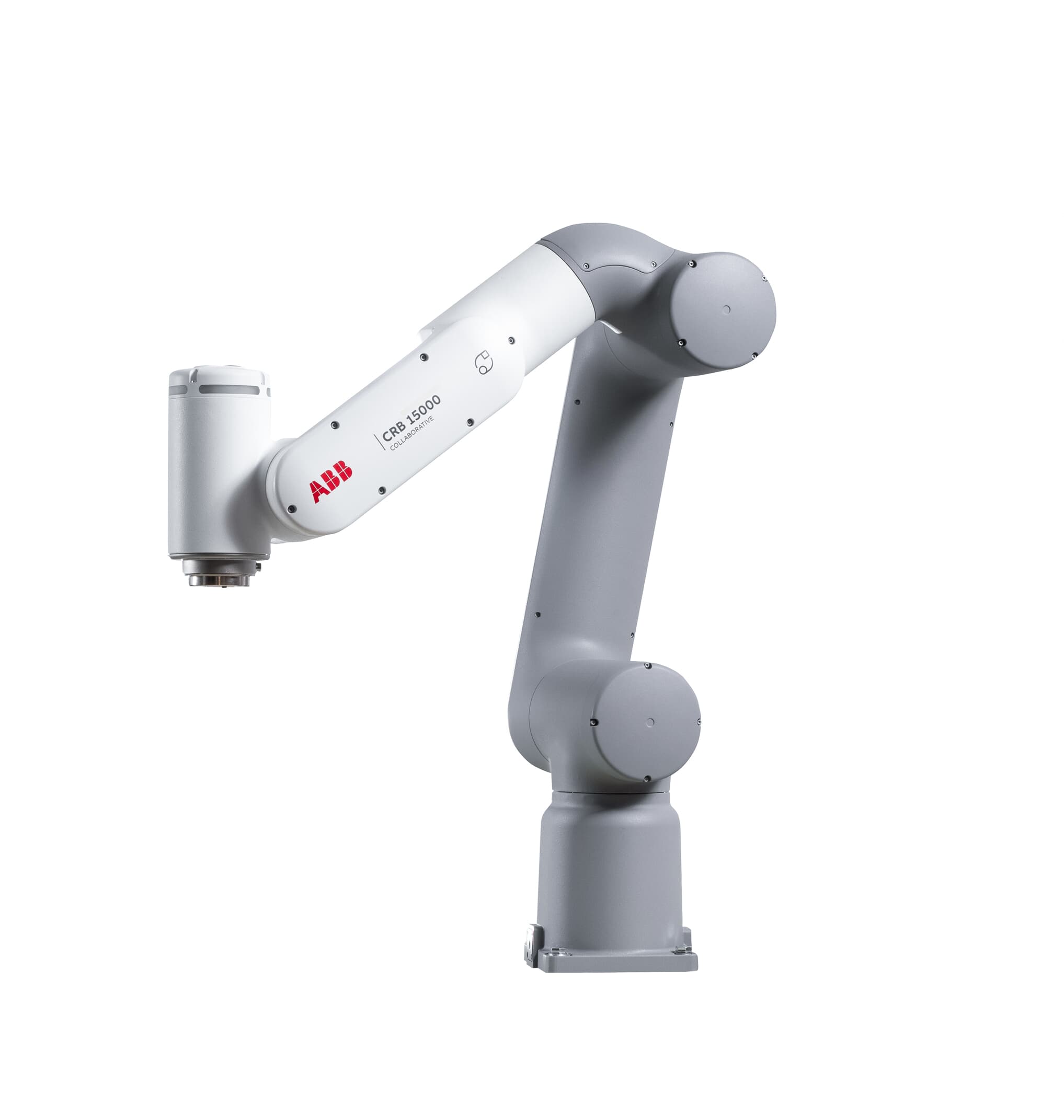}}% 
        \subcaption{\label{fig:gofa_photo}}%
    \end{subfigure}%
    \caption{
    Industrial cuspidal robots.
    (a)~FANUC CRX-10iA/L.
    (b)~Kinova Link 6.
    (c)~ABB GoFa CRB 15000 5 kg.
    }
    \label{fig:robot_photos}
\end{figure}

Recently, cobots such as the FANUC CRX-10iA/L~\cite{crx} (Fig.~\ref{fig:CRX_photo}) and Kinova Link 6~\cite{link6} (Fig.~\ref{fig:kinova_photo}) have come to market. These robots have an offset wrist and do not have three parallel axes. This means IK cannot be solved only with quadratics, and there may be more than eight IK solutions. Practitioners have pointed out notable difficulties with path planning and IK for these robots~\cite{wenger2023review}. The reason for this difficulty is because robots such as the CRX and Link 6 are cuspidal.

Cuspidal robots can travel between IK solutions without encountering a singularity. They have IK that cannot be solved with quadratics and have certain constraints on kinematic parameters such as link lengths.
For a long time, this property was thought to be impossible\cut{, and a}\new{. A}n incorrect proof of this impossibility was published~\cite{borrel1986study}, but this proof was later proven wrong~\cite{wenger1992new}.
Even today, robot cuspidality is rarely considered~\cite{wenger2023review} but a robot with a randomly chosen geometry is almost surely cuspidal~\cite{salunkhe2024kinematic}.
It appears \new{manufacturers and practitioners are building and using} cuspidal robots \cut{are being used} without \cut{manufacturers or practitioners }realizing they are cuspidal or fully appreciating the resulting path planning challenges.

\new{Although there is limited research in this field, there has been a recent resurgence in interest because cuspidal manipulators are becoming commercially widespread for the first time.
Most research discusses issues arising from manipulators being cuspidal, and many papers conclude cuspidal robots should be avoided. For example, \cite{salunkhe2023cuspidal} argued cuspidal robots should be avoided in collaborative applications, and \cite{wenger2023review} argued it is reasonable to avoid designing cuspidal robots.}

The following example illustrates a case where cuspidality can cause path planning issues. Consider the case of robotic welding, which is an advertised use case for many industrial cuspidal robots. A robot operator may program a closed path for a welding operation, and there may be no apparent problem because the robot can perform the entire weld. However, since the robot is cuspidal, the robot can perform a nonsingular change of solution during the welding operation. This means when the robot executes the weld for a second time, the robot may get stuck at a singularity\cut{, or}\new{. Moreover}, if the first command in the program is a joint-motion command, the robot may produce unexpected large motion to return to the original configuration!

For practitioners who are only used to non-cuspidal robots, their existence and many of their properties are nonintuitive.
It is difficult to give meaningful names to IK solutions (e.g., elbow up / down) because a robot may move between these solutions with a nonsingular path (meaning they are in the same \textit{aspect}).
Furthermore, there may be smooth task-space paths such as straight-line Cartesian \texttt{MoveL} commands with IK solutions entirely in one aspect that cannot be smoothly followed.
    This is because local boundary singularities prevent end effector movement within the workspace. In general, these singularities form surfaces that cannot be avoided with small path deviations, unlike, say, a wrist singularity in a conventional industrial robot with a spherical wrist.
In some cases, a closed Cartesian path can be followed once, but attempting to follow the path a second time results in the robot getting stuck at a singularity because following the path causes the robot to switch IK solutions.

Cuspidal robots highlight the importance of considering all IK solutions during path planning and demonstrate the need for IK solvers which can find all IK solutions.
\new{Missing IK solutions is problematic because feasible paths may be missed, resulting in a failure to connect two points or picking a suboptimal path.} 
(In fact, finding all IK solutions is important for path planning even for noncuspidal robots as certain paths may be non-optimal or even infeasible depending on the choice of initial IK solution.) Many IK solvers are Jacobian-based iterative solvers, and these do not return all IK solutions.

IK solvers used by industrial robots which return multiple solutions may have problems. ROBOGUIDE, the software used by FANUC to control their CRX line of robots, only provides up to eight IK solutions even though the CRX has up to sixteen in general~\cite{salunkhe2024kinematic}. Furthermore, \new{ROBOGUIDE labels} IK solutions \cut{in ROBOGUIDE are labeled} using shoulder/elbow/wrist configuration parameters, but these descriptions may not correctly match the configuration of the arm and may not change when changing IK solutions. In fact, we can find end effector poses for the CRX with sixteen IK solutions but only two aspects~\cite{salunkhe2023cuspidal}\cut{!}\new{.} In Section~\ref{sec:path_planning_challenges}, we show very similar issues for the ABB GoFa (Fig.~\ref{fig:gofa_photo}) for the first time.

In this paper, we use our IK solver IK-Geo \cite{elias2025_6dof} to efficiently find all IK solutions. IK-Geo also finds continuous approximate solutions to provide robust IK solutions at singularities. The Husty-Pfurner~IK algorithm \cite{husty2007new} is a common choice in literature because it performs well on generic robots with no simplifying cases of intersecting or parallel axes. However, all industrial robots have some intersecting or parallel axes, and IK-Geo efficiently solves the IK problem by exploiting these simplifications.

\new{
By leveraging the efficient and complete IK solutions from IK-Geo, the goal of this paper is to identify cuspidal robots and perform path planning and path optimization for these robots.
Path planning algorithms for noncuspidal robots may not work correctly for cuspidal robots because they
    may not consider all IK solutions,
    may not check for changes of solutions for closed task-space paths, 
    or may check for connectivity based on aspect or the presence of IK solutions throughout the path.
Although there exist algorithms for noncuspidal robots for optimizing the rigid-body transformation of a task-space path, no optimization algorithms have been proposed specifically for cuspidal robots before now. }

The contributions of this paper are the following:
\begin{itemize}
    \item We demonstrate identifying cuspidal robots using IK-Geo. We show the ABB GoFa is cuspidal for the first time. We also show robots with three parallel axes may be cuspidal for the first time.
    \item We develop a joint-space path planning framework compatible with cuspidal robots by reducing the problem to a graph search. For a given task-space path, the path planner finds all feasible joint-space paths represented as a weighted directed graph. The optimal joint-space path is found by solving the shortest path problem.
    \item We demonstrate path optimization using the graph-based path planning framework. For a given task-space path, we efficiently find the optimal rigid-body transformation of the entire path to minimize a user-defined optimization metric while ensuring feasibility.
\end{itemize}

The remainder of this paper is structured as follows. We
\new{explain our notation for robot kinematics,}
provide definitions, give an overview of cuspidal path planning challenges, and discuss related literature in Section~\ref{sec:background}.
We identify cuspidal robots in in Section~\ref{sec:cuspidal_ID}.
We develop and demonstrate path planning method in Section~\ref{sec:path_planning} and use it for path optimization  in Section~\ref{sec:path_opt}. We conclude in Section~\ref{sec:conclusion}.

Code is available in a publicly accessible repository.\footnote{\url{https://github.com/rpiRobotics/cuspidal-path-planning}}

\section{Background}\label{sec:background}
\begin{figure}[t]%
    \begin{subfigure}[t]{0.6\linewidth}%
        \centering%
        \includegraphics[scale=0.5, clip]{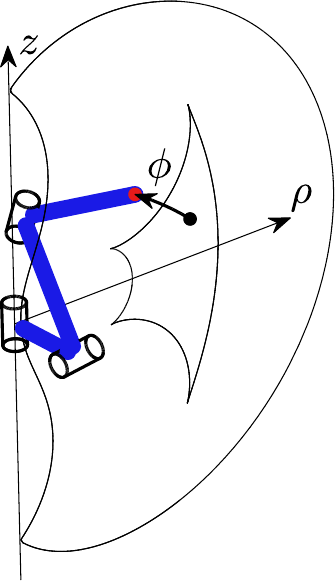}%
    \subcaption{\label{fig:cupsidal_3R_3d}}%
    \end{subfigure}%
    \begin{subfigure}[t]{0.4\linewidth}%
        \includegraphics[scale=0.5, clip]{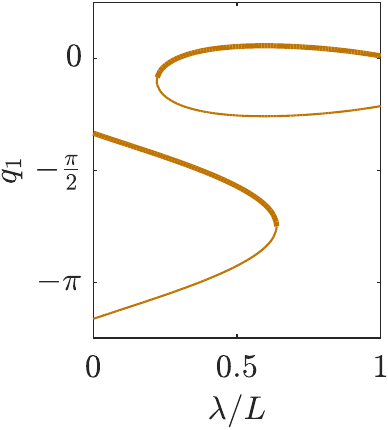}%
    \subcaption{\label{fig:3R_infeasible}}%
    \end{subfigure}\\[1em]
    \begin{subfigure}[t]{0.6\linewidth}%
        \includegraphics[scale = 0.48, clip]{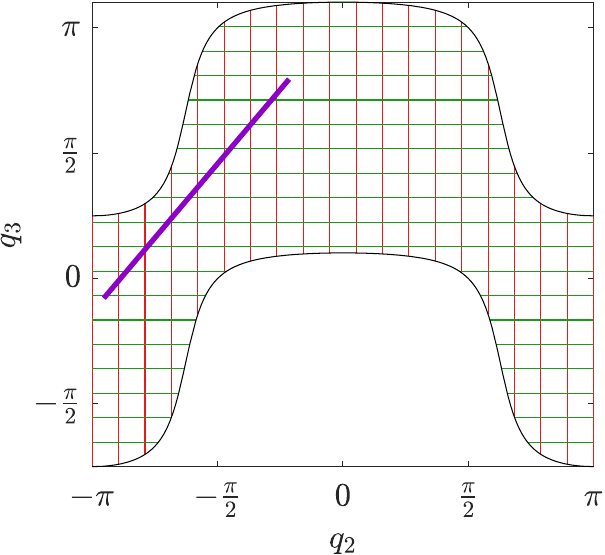}%
    \subcaption{\label{fig:joint_space_grid}}%
    \end{subfigure}%
    \begin{subfigure}[t]{0.4\linewidth}%
        \includegraphics[scale = 0.48, clip]{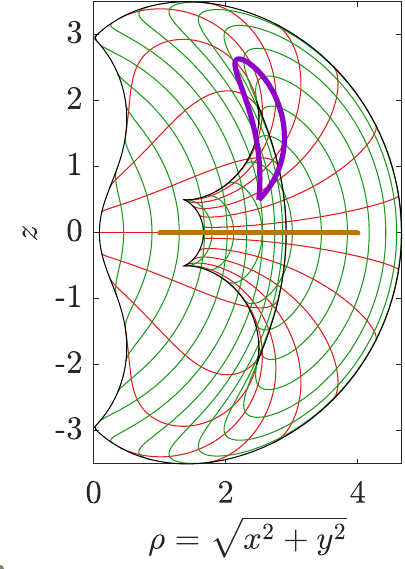}%
    \subcaption{\label{fig:task_space_grid}}%
    \end{subfigure}%
\caption{
Cuspidal 3R manipulator with an infeasible path (straight line in task space) and a nonsingular change of solution (straight line in joint space).
(a)~3D view \new{using cylindrical coordinates \((\rho, \phi, z)\)}. Points with singularities are marked for the \(\phi=0\), \(\rho>0\) half-plane.
(b)~All solutions for joint angle 1 over the infeasible path. Line width corresponds to which aspect the IK solution is in.
(c)~One aspect in joint space bounded by singularities. Nonsingular change of solution is shown.
(d)~Forward kinematics mapping of one aspect to task space. Also shown are the infeasible path and the nonsingular change of solution.}
    \label{fig:cuspidal_topology}
\end{figure}

\new{

The configuration of a serial revolute robot manipulator is determined by the joint angle vector
\begin{equation}
    q = [q_1 \ q_2 \  \dots \ q_6]\tr \in \mathbb{R}^6,
\end{equation}
where in this case the robot has six degrees of freedom. A 6-DOF revolute manipulator is denoted 6R, and a 3-DOF revolute manipulator is denoted 3R. We will consider nonredundant manipulators in this paper, meaning the joint space and task space have the same dimension.

If we consider joint angles~\(q_i\) and~\(q_i+ 2\pi\) equivalent, such as when there are no joint limits, then we can model the joint space as a torus, that is, \(q \in S^6\) and~\(q_i \in [-\pi, \pi]\).

Forward kinematics \(f_{0T}\) is the mapping from joint space to task space:
\begin{equation}
    \chi_{0T} = (R_{0T}, p_{0T}) = f_{0T}(q),
\end{equation}
where \(\chi_{0T} \in \mathrm{SE}(3)\) is the pose of the task (end effector) frame in the robot base frame. This consists of the end effector orientation
\(R_{0T} \in \mathrm{SO}(3)\) and Cartesian end effector position~\(p_{0T} \in \mathbb{R}^3\). For a 3-DOF robot, the end effector pose only consists of the position. In addition to using Cartesian coordinates \((x,y,z)\), we may also represent the end effector position using cylindrical coordinates~\((\rho, \phi, z)\), where \(\rho = (x^2 + y^2)^{1/2}\) and \(\phi = \mathrm{ATAN2}(y, x)\).

Inverse kinematics (IK) is the mapping from a given end effector pose to its preimage, which is the set of all joint angle vectors achieving that end effector pose:
\begin{equation}
f_{0T}^{-1}(\chi_{0T}) = \{ q \in \mathbb{R}^6 \mid f_{0T}(q) = \chi_{0T} \}.
\end{equation}
There may be multiple IK solutions or even a continuum of solutions for a given end effector pose.

The kinematic Jacobian~\(J\), which is a function of~\(q\), defines the linear mapping from joint velocity to task velocity:
\begin{equation}
    \nu_{0T} = [\mathscr v_{0T} \ \omega_{0T}]\tr = J(q) \dot q,
\end{equation}
where~\(\nu_{0T} \in \mathbb{R}^6 \cong \mathfrak{se}(3)\)
is the end effector spatial velocity composed of the linear velocity~\(\mathscr{v}_{0T} \in \mathbb{R}^3\)
and the angular velocity~\(\omega_{0T} \in \mathbb{R}^3\). For a 3-DOF robot, the end effector velocity is just the linear velocity.

A kinematic singularity occurs when~\(J(q)\) loses rank, which occurs when~\(\det(J) = 0\). At a singular configuration, there is some spacial velocity direction which cannot be achieved, and there is some direction in joint space which results in zero spatial velocity (perhaps only instantaneously). This internal robot motion without moving the end effector is called self-motion.

A task-space path~\(\underline \chi _{0T}(\lambda)\) is a mapping from a path variable~\(\lambda \in \mathbb{R}\) to an end effector pose. A joint-space path~\(\underline q(\lambda)\) is defined similarly.
Whereas paths are indexed by the path variable~\(\lambda\), trajectories are indexed by time~\(t\).
}

\subsection{Definitions}

The following definitions apply to nonredundant serial manipulators.

\paragraph*{Cuspidal manipulator}
A cuspidal manipulator can move between unique IK solutions without encountering a singularity. In other words, a manipulator is cuspidal if it has several IK solutions in an aspect \cite{wenger2023review}. 

\paragraph*{Aspect} An aspect is a largest connected region in the robot joint space free of singularities \cite{wenger2023review}. Aspects are bounded not just by singularities but also by joint limits when they exist.  Joint limits can convert a cuspidal robot into a noncuspidal robot \cite{wenger2023review}.

\paragraph*{Characteristic surface}
Points in the characteristic surface of an aspect (in joint space) correspond to end effector poses that have a singular solution bounding same aspect. Noncuspidal robots have no characteristic surfaces.

\paragraph*{Feasible path} A workspace path is feasible if it corresponds to a continuous joint-space path. Most other literature only considers singularity-free paths, but a path can be continuous and pass through a singularity.

\paragraph*{Repeatable path} A closed workspace path is repeatable if it is feasible and can be repeated arbitrarily many times. A nonrepeatible path can only be followed a finite number of times. A \textit{regular} repeatable path returns to the same IK solution after each task-space cycle. Non-regular repeatable paths have a different cycle period and switch IK solutions during each workspace cycle. In noncuspidal robots, if a closed path is feasible then it is also repeatable. For cuspidal 3R robots, nonsingular changes of solutions are necessarily nonrepeatable. For cuspidal 6R robots, there are repeatable paths that are not regular~\cite{salunkhe2023cuspidal}.

\phantom{} % Make sure there is a break between the above and below paragraphs

Parallel robots can also be cuspidal \cite{wenger2023review, macho2012planning, coste2012simple, bamberger2008assembly, manubens2012cusp}, but the roles of the forward and inverse kinematics mapping are reversed. Cuspidal parallel robots have closed active-joint-space paths with different beginning and ending end effector poses.

\subsection{Cuspidal Robot Path Planning Challenge}\label{sec:path_planning_challenges}
It is imperative that practitioners know when a robot is cuspidal and how the cuspidality can make path planning more challenging. For example, the ABB IRB 6400C was a cuspidal manipulator that was pulled from commercialization one year after being launched in 1996 perhaps because of path planning difficulties \cite{wenger2007cuspidal}.

The workspace of cuspidal manipulators is divided into regions of different numbers of IK solutions. At the boundary of these regions, multiple IK solutions converge and terminate at a boundary singularity.
However, in the case of cuspidal robots, a single aspect can have IK solutions on both sides of a boundary. Some path planners may cause a sudden jump in joint angles if the IK solution switches from one of the terminating solutions to the other solution in the same aspect~\cite{salunkhe2023trajectory}. 
Sometimes, choosing a different initial IK solution makes the path feasible. Other times, no initial IK solution works.

Graphically, robot cuspidality can be understood by considering the forward kinematics mapping of one aspect in joint space to its image in task space. The product of exponentials kinematic parameters for the canonical cuspidal 3R manipulator (Fig.~\ref{fig:cuspidal_topology}), which was perhaps first described in \cite{wenger1992new}, are 
\begin{equation}
\begin{gathered}
    p_{01} = 0,\ 
    p_{12} = e_x,\
    p_{23} = 2 e_x + e_y,\
    p_{3T} = 1.5 e_x,\\
    h_1 = h_3 = e_z,\
    h_2 = e_y.
\end{gathered}
\end{equation}
In the product of exponentials convention (which is further described in \cite{elias2025_6dof}), \(p_{01}\) is the position of the first link frame in the base frame, \(p_{i-1,i}\) is the position of link frame \(i\) in the \(i-1\) frame, and \(p_{i,T}\) is the position of the end effector task frame in the \(i\) frame. \cutmath{\(h_i\) is t}\new{T}he joint axis direction\cut{s} for joint \(i\) represented in the~\(i-1\) frame \new{is denoted by \(h_i\)}. \cutmath{\(e_x\), \(e_y\), and \(e_z\) are t}\new{T}he Cartesian unit vectors \new{are denoted by \(e_x\), \(e_y\), and \(e_z\)}.

In Fig.~\ref{fig:joint_space_grid} and \ref{fig:task_space_grid}, we see the mapping of one aspect overlaps with itself in the center, meaning there are two IK solutions for a single end effector pose within this aspect. This permits a nonsingular path between two IK solutions. A 3R serial manipulator is cuspidal if and only if it has a cusp point in the forward kinematics mapping of the locus of singular points. Graphically, the cusp point is the point about which the grid lines change directions to allow the mapping to overlap with itself. When crossing between task-space regions with different numbers of IK solutions, the joint-space path may be feasible only for certain IK solutions. For a task-space path with multiple crossings, there may be no feasible joint-space path regardless of the initial IK solution. We show such an infeasible path in Fig.~\ref{fig:3R_infeasible} and \ref{fig:task_space_grid}, where a straight-line path passing from the two-solution region to the four-solution region and back to the two-solution region has no feasible paths. Interestingly, we can find an arbitrarily short infeasible path by moving it closer to a cusp point.

Although plotting the forward kinematic mapping is straightforward for 3R robots and 6R robots with spherical wrists, this becomes more challenging for other robots with singularities that depend on the end effector orientation.

In commonly-used industrial robot software, when the robot encounters a singularity when performing a primitive task-space move such as a straight-line Cartesian move (\texttt{MoveL}), there is no way to determine which of the following two cases is true:
\begin{enumerate}
    \item The path is feasible but the wrong initial IK solution was picked, or
    \item The path is infeasible regardless of the choice of initial IK solution.
\end{enumerate}

We demonstrate cuspidal path planning issues for the ABB GoFa in RobotStudio and the FANUC CRX-10iA/L in ROBOGUIDE. \new{These are two newer robots from large manufacturers. This means the simulators are highly realistic, as they use virtual controllers which match the behavior of the real robot. This also means there is a practical application since these robots are currently in production. As neither robot has three intersecting or parallel joint axes, their IK is not solved in closed form, making them both interesting and challenging examples. }

\begin{figure}[t]
    \centering
    \raisebox{-0.5\height}{\includegraphics[scale=0.5, clip]{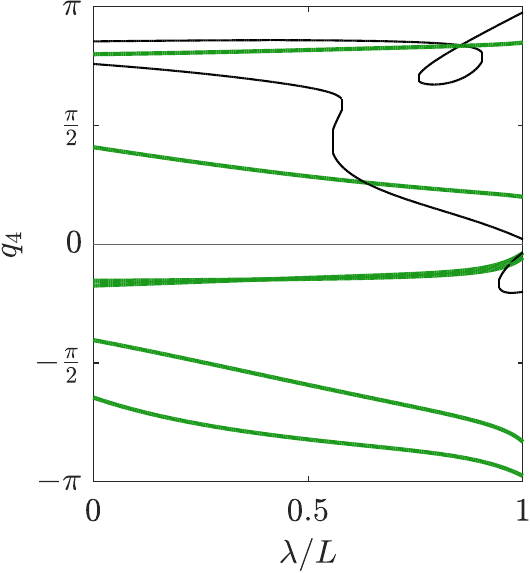}}%
    \raisebox{-0.4\height}{\includegraphics[width=1.5in]{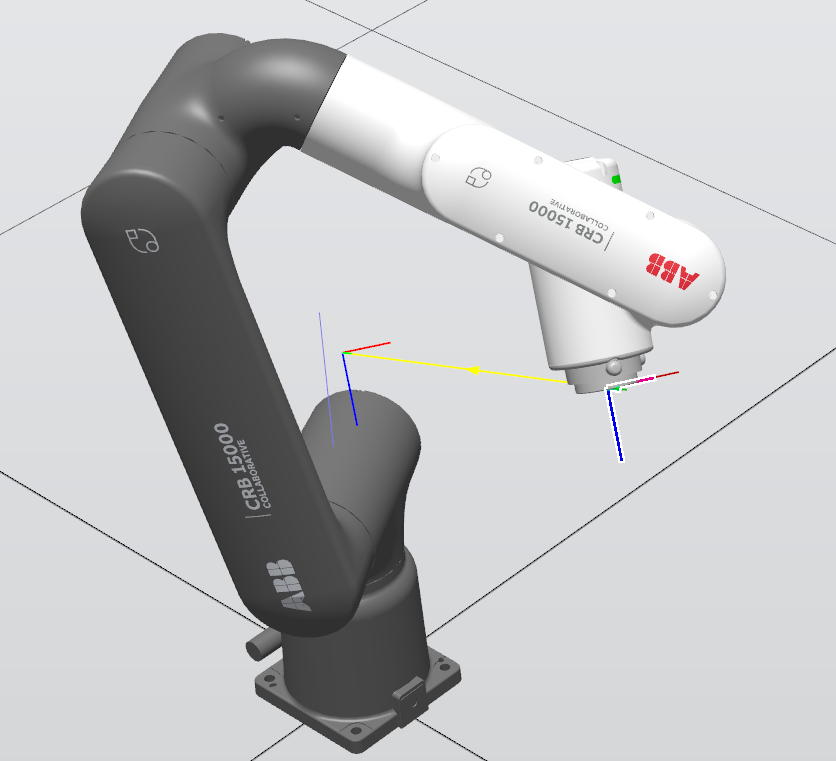}}
    \caption{\texttt{MoveL} path for the ABB GoFa which is feasible depending on the initial pose. In this case, there are eight initial and ten final IK solutions but only six feasible paths.}
    \label{fig:gofa_MoveL_issues}
\end{figure}

For the ABB GoFa, it is easy to find examples of \texttt{MoveL} paths that are entirely infeasible or feasible only for certain choices of initial IK solution (Fig.~\ref{fig:gofa_MoveL_issues}).
For feasible paths, the configuration label may change between the start and end, so feasibility cannot be determined simply by comparing the configuration labels at the beginning and end of the path. During motion, the \texttt{ConfL \textbackslash{}Off} command must sometimes be used so the robot controller allows passing between configuration parameters, even though the robot does not encounter a singularity.
This is the first time cuspidal behavior has been shown for the GoFa in the literature.

We also raise the concern for the first time that RobotStudio does not find all IK solutions for the GoFa. The robot has up to 16 solutions, but \new{RobotStudio returns }only up to 8 solutions\cut{ are returned}. That the solutions are labeled from 0 to 7  is explicitly called out for the GoFa in the programming manual \cite[Sec. 3.16]{rapid_ref_manual}.

\begin{figure}[t]
    \centering
    \raisebox{-0.5\height}
    {\includegraphics[scale=0.5, clip]{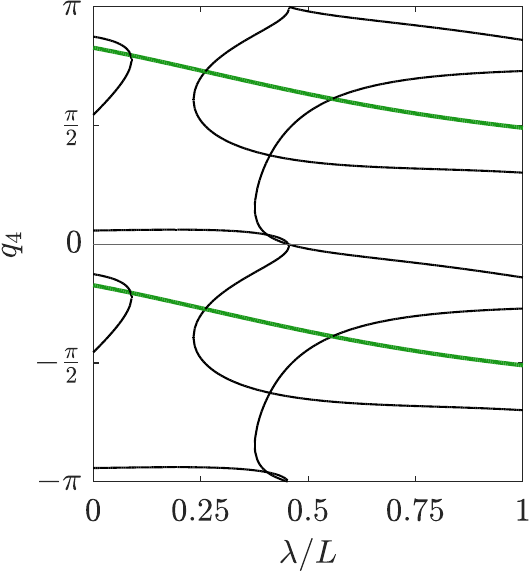}}%
    \raisebox{-0.4\height}{\includegraphics[width=1.5in]{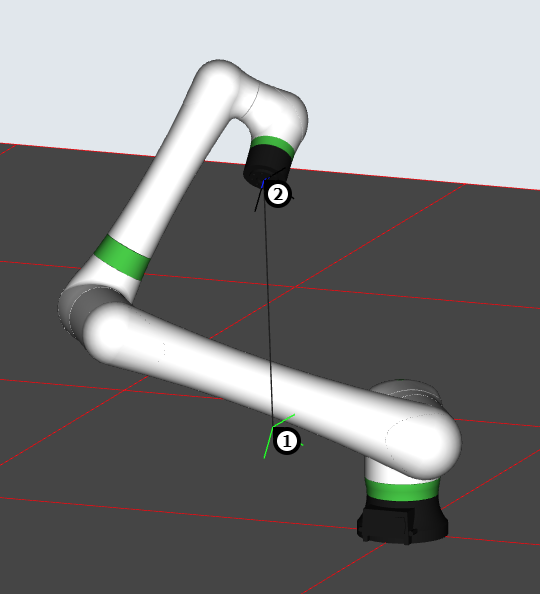}}
    \caption{\texttt{MoveL} path for the FANUC CRX-10iA/L with eight initial and final IK solutions, up to twelve intermediate solutions, and only two feasible paths.}
    \label{fig:crx_MoveL_issues}
\end{figure}

A similar example of a challenging path for the CRX-10iA/L is shown in Fig.~\ref{fig:crx_MoveL_issues}. In searching the workspace for random \texttt{MoveL} paths, it appears the CRX is more likely than the GoFa to have paths that are feasible depending on the choice of IK solution or entirely infeasible paths.

\begin{figure}[t]
    \centering
    \includegraphics[scale=0.5, clip]{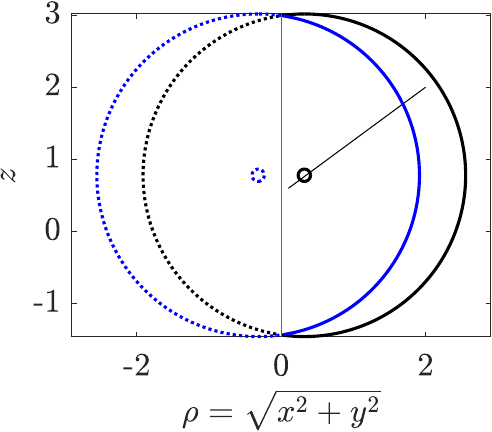}
    \includegraphics[width=1.5in]{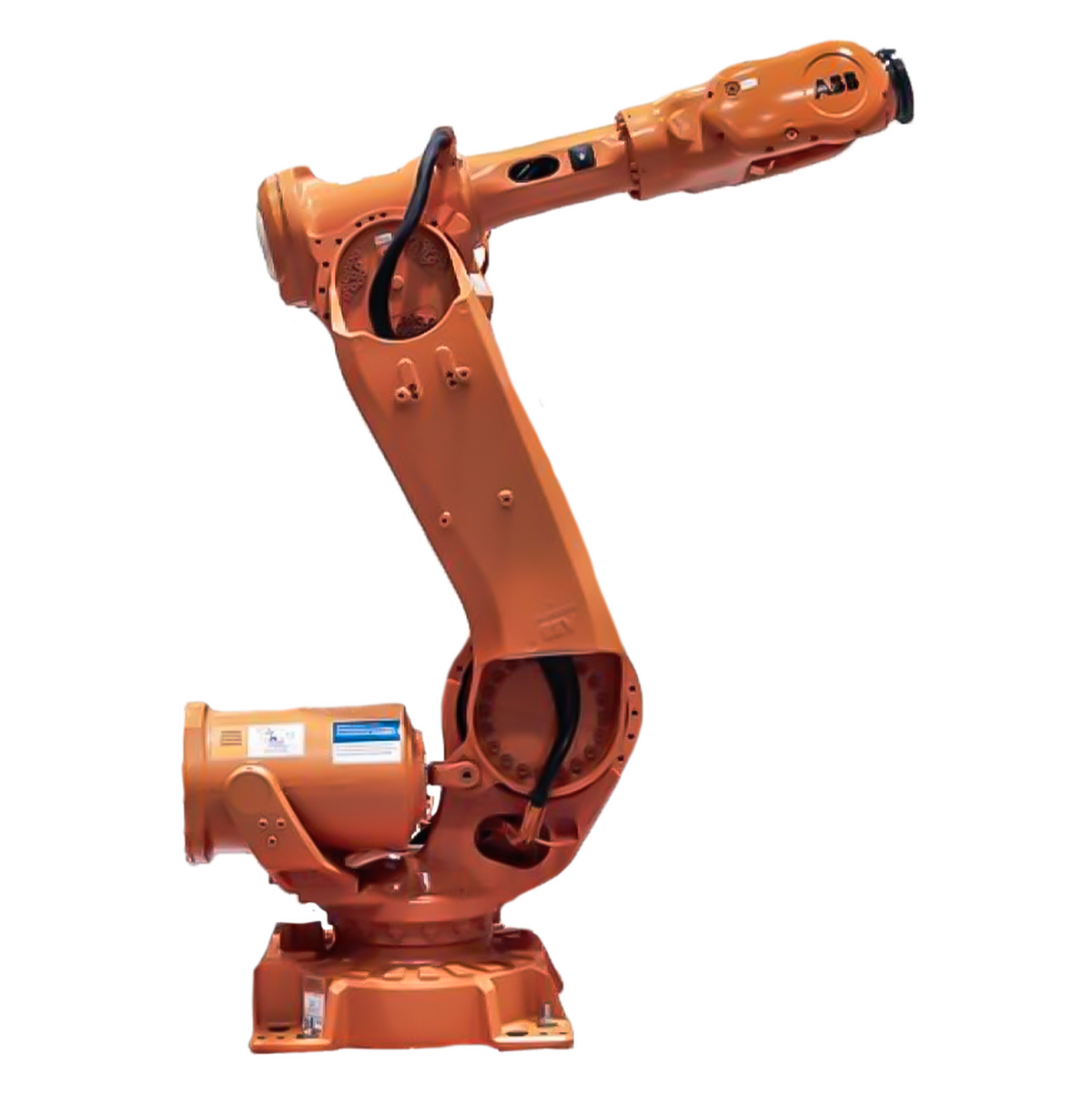}
    \caption{Path for the noncuspidal ABB IRB 6640 which is entirely in the workspace but is infeasible. The blue annulus centered at \(\rho<0\) shows the task space reachable by shoulder back configurations. The black annulus centered at \(\rho>0\) shows the task space reachable by shoulder forward configurations. The path shown passes through the hole in the black annulus and also leaves the outer boundary of the blue annulus. Therefore, neither configuration can achieve the entirety of the desired path.}
    \label{fig:6640_infeasible}
\end{figure}

There are similar path planning challenges in noncuspidal robots which perhaps have not been fully appreciated in the literature or by practitioners.
Even for noncuspidal robots, having the path entirely in the workspace does not guarantee the path is feasible. This is because each aspect may have a different shape in task space. Any path that passes through both a region covered only by one aspect and a region only covered by another aspect will be infeasible for those IK solutions. An example of a path for the ABB IRB 6640 which is entirely in the workspace but is infeasible is shown in Fig.~\ref{fig:6640_infeasible}. (In practice, the joint limits prevent this from being a problem.) Similarly, noncuspidal robots may have paths that are feasible only for certain initial IK solutions.

Even for noncuspidal manipulators, labeling the configuration is difficult in some cases. For example, robots with three intersecting or parallel axes but no other simplifications have their IK solved with a quartic equation, and depending on the remaining kinematics parameters, they may be noncuspidal. In these cases, labels given to each aspect may have no clear geometric meaning to the robot operator.

\subsection{Related Literature}
\cutmath{Cuspidal manipulators have been known for some time. Although there is limited research in this field, there has been a recent resurgence in interest because cuspidal manipulators are becoming commercially widespread for the first time.

Most research discusses issues arising from manipulators being cuspidal, and many papers conclude cuspidal robots should be avoided. For example, \cite{salunkhe2023cuspidal} argued cuspidal robots should be avoided in collaborative applications, and \cite{wenger2023review} argued it is reasonable to avoid designing cuspidal robots.}

There have been some important recent developments in the identification and path planning for cuspidal manipulators. There is also some important precedent for path optimization, but only for noncuspidal manipulators.

\subsubsection{Identification of Cuspidal Manipulators}
There are several ways to prove that a serial robot is cuspidal or noncuspidal.

If the robot can be solved without using polynomials of degree three or higher, then the robot is not cuspidal, or, equivalently, if the determinant\new{~\(\det(J)\)} factors into three terms, then the robot is noncuspidal. A list of noncuspidal 6R robots with this property is found in~\cite{salunkhe2023cuspidal}. \cut{There are m}\new{M}any kinematic simplifications \cut{that can }result in a polynomial of degree two~\cite{mavroidis1994structural}. Two common cases for revolute manipulators to be solved with only quadratics is to have a spherical wrist or three parallel axes plus two other intersecting or parallel axes~\cite{elias2025_6dof}. If a 6R robot has a spherical wrist, then it is cuspidal if and only if the 3R robot formed by the first three links is cuspidal.

A 3R robot is cuspidal if and only if it has a cusp point in the locus of singular points in the workspace, which is equivalently a triple point with exactly three solutions.
The IK problem for a 3R manipulator (Subproblem~5 in~\cite{elias2025_6dof}) can be solved as the intersection of two ellipses, and the singularities correspond to tangencies between ellipses~\cite{thomas2016distance}. A cusp point corresponds to osculating contact between two ellipses.
Necessary and sufficient kinematic conditions for cuspidality have been found for 3R orthogonal manipulators~\cite{wenger2023review}. \new{Researchers have}\cut{It has} only \cut{been} recently proven that a general 3R manipulator is cuspidal if and only if it has a cusp point~\cite{salunkhe2022necessary}. 

It is unknown if 6R manipulators are cuspidal if and only if they have a cusp point, and this condition is false for parallel manipulators.
In parallel manipulators, a nonsingular change of solution does not require encircling a cusp point and can instead occur due to encircling an \(\alpha\)-curve~\cite{bamberger2008assembly, thomas2016distance}.
\cut{It is unfortunate that}\new{Unfortunately,} the name ``cuspidal" appears to be a misnomer as cusp points are not necessary in general for a manipulator to be cuspidal. 

Even 2-DOF manipulators can be cuspidal. In particular, planar 2-X manipulators can be cuspidal~\cite{furet2019kinetostatic}. These are tensegrity structures with two X-mechanisms, which have variable instantaneous centers of rotation, connected in series.

Cuspidality can be proven or disproven by analyzing singularities. For example, the Kinova Link 6 has been proven to be cuspidal by identifying that all singularities require at least two specific joint angles~\cite{asgari2024singularity}.

A certified algorithm for deciding cuspidality of serial revolute manipulators based on real algebraic geometry was proposed in~\cite{chablat2022deciding}. However, this method is computationally complex, very hard to implement, not currently implemented for 6R robots, and not able to handle collision constraints~\cite{salunkhe2023cuspidal}.

\cutmath{The identification of 6R manipulators has been reduced to the path planning problem of finding a nonsingular path between IK solutions, which we discuss further in Section~\ref{sec:path_planning_related_work}.
\cite{salunkhe2023trajectory}~identified cuspidal robots by finding a linear joint-space \texttt{MoveJ} path between IK solutions and then performing path optimization to avoid passing through a singularity twice.
\cite{salunkhe2024kinematic}~identified robots by solving the optimal path planning problem with a multiple shooting approach.
Finding a nonsingular change of IK solution proves a robot is cuspidal. Not finding such a path means a robot is likely noncuspidal, but this is not proof.}

\new{The problem of identifying if a 6R manipulator is cuspidal has been reduced to a path planning problem: A robot is cuspidal if a path planner can find a nonsingular path between two different IK solutions of the same end effector pose. We discuss the path planning problem further in Section~\ref{sec:path_planning_related_work}.
One proposed identification method~\cite{salunkhe2023trajectory} used linear joint-space \texttt{MoveJ} paths as the initial guess for a path optimization algorithm that adjusted the path to avoid singularities.
A subsequent study \cite{salunkhe2024kinematic}~identified cuspidal robots by formulating the path planning problem as an optimization problem and solving it with a multiple shooting approach.
If a nonsingular path between IK solutions is found, this proves the robot is cuspidal. Conversely, failing to find such a path suggests the robot is likely noncuspidal, though this does not constitute a formal proof.}

\subsubsection{Path Planning for Cuspidal Manipulators}\label{sec:path_planning_related_work}

Path planning for cuspidal robots is not well studied \cite{salunkhe2023cuspidal}.  There are several different path planning problems for cuspidal manipulators.

The point-to-point path planning problem prescribes the starting and ending pose of the robot but not the intermediate path of the end effector. A special case is to find a feasible closed path between different IK solutions. Finding such a path which is nonsingular proves the robot is cuspidal.

A singularity-free time-optimal point-to-point trajectory planning algorithm for cuspidal robots was presented in~\cite{marauli2023time}. The algorithm minimizes the weighted sum of terminal time and the \(L_2\) norm of jerk (for regularization) subject to constraints on joint angles, velocities, accelerations, jerk, and torque, as well as to a constraint on distance to singularities. The solution method uses a multiple shooting approach for a given pair of initial and final IK solution candidates with the same \(\sign(\det(J))\). The optimization is done for all candidate pairs, also accounting for possible \(\pm 2\pi\) offsets by considering the interval \([-2\pi,\ 2\pi]\) for each joint angle. Demonstrations were presented for the canonical cuspidal 3R manipulator and the FANUC CRX-10iA/L.

A different path planning problem is finding a feasible joint-space path for a given task-space path. This can be for open paths or closed paths.
\new{This is the problem we will solve in Section~\ref{sec:path_planning}.}
To the best of our knowledge, there has only been one path planning algorithm proposed for cuspidal robots following a prescribed task-space path presented in \cite{salunkhe2024kinematic, salunkhe2023cuspidal} (the algorithms, presented as flow charts, are slightly different between \cite{salunkhe2024kinematic} and \cite{salunkhe2023cuspidal} in the case that the path doesn't cross a characteristic surface.) The algorithm is:

\begin{enumerate}
    \item Pick one IK solution along the path with the smallest number of IK solutions.
    \item If the path crosses a characteristic surface, check the connectivity of the point to the initial and final IK solutions. If the path is closed, also check if the path is regular, or if it exhibits a nonsingular change of solution in which case it needs to be determined if the path is repeatable. 
    \item If the path does not cross a characteristic surface, check if the path is continuous.
\end{enumerate}
The algorithm description does not specify if all possible initial and final IK solutions should be evaluated.

Although a similar problem is solved in this paper, there are several key differences between the algorithm presented in \cite{salunkhe2024kinematic, salunkhe2023cuspidal} and our proposed algorithm:
\begin{itemize}
    \item We evaluate not just the feasibility but also the cost of each possible joint-space path.
    \item We reduce the problem to a graph search over a weighted directed acyclic graph (DAG). There are many existing software tools to efficiently perform this search.
    \item We demonstrate incorporating least-squares and continuous approximate IK solutions. Although our algorithm can avoid singularities, it also allows for paths that go through singularities.
    \item The path planning algorithm is fast enough to be used inside an optimization loop thanks to the computational efficiency of IK-Geo.
    \item Whereas the algorithm in \cite{salunkhe2024kinematic, salunkhe2023cuspidal} is presented using a high-level description, our algorithm is presented using a particular graph-based implementation with publicly available code examples.
\end{itemize}

The Descartes path planning algorithm~\cite{de2017cartesian} is part of ROS industrial and also solves the path planning problem using a graph method. However, this algorithm does not account for problems specific to cuspidal manipulators.

\subsubsection{Path Optimization for Cuspidal Manipulators}
\cutmath{Although there exist algorithms for noncuspidal robots for optimizing the rigid-body transformation of a task-space path, no optimization algorithms have been proposed specifically for cuspidal robots before now. Path planning algorithms for noncuspidal robots may not work correctly for cuspidal robots because they
    may not consider all IK solutions,
    may not check for changes of solutions for closed task-space paths, 
    or may check for connectivity based on aspect or the presence of IK solutions throughout the path.}

Proposed path-planning algorithms for non-cuspidal robots are often related to machining tasks such as drilling or milling where robot stiffness must be optimized. These tasks are commonly achieved using robots with spherical wrists \cite{weingartshofer2021optimal, ye2021task, dumas2012workpiece, liao2021optimization, he2024fast} or with a UR robot \cite{balci2023optimal, stradovnik2024workpiece}. Optimization was done for a Stewart platform in~\cite{lopes2011optimization}.
Some literature resolves the redundant degree of freedom about a spinning tool \cite{dumas2012workpiece, liao2021optimization}. In~\cite{he2024fast}, the relative path of two robots was optimized, which required resolving six redundant degrees of freedom.
Many algorithms use\new{d} optimization strategies like particle swarm optimization (PSO)~\cite{ye2021task, liao2021optimization} or genetic algorithms~\cite{lopes2011optimization}.
Objective functions included optimizing joint velocities~\cite{stradovnik2024workpiece, weingartshofer2021optimal}, cycle time~\cite{he2024fast},
joint torque and manipulability~\cite{balci2023optimal},
stiffness ~\cite{liao2021optimization, lopes2011optimization},
contour accuracy (by considering cutting forces and stiffness)~\cite{ye2021task, dumas2012workpiece}, and power consumed~\cite{lopes2011optimization}.
Tool offset was optimized in addition to base position in \cite{weingartshofer2021optimal}. Furthermore, \cite{weingartshofer2021optimal}~proposed a path planner similar to ours (it is implicitly a graph search) that can go through singularities.
Rather than optimizing over six degrees of freedom, \cite{stradovnik2024workpiece} optimized only over the three degrees of freedom in the placement of a workpiece on a table.
Euler angles were used for parameterization in \cite{lopes2011optimization, liao2021optimization}, but the \new{representation} singularity problem was not addressed.

\section{Cuspidal Robot Identification}\label{sec:cuspidal_ID}
In this section, we describe an efficient method to determine if a robot is cuspidal. Using this general method, we show the ABB GoFa and a robot with three parallel joint axes are cuspidal.

\subsection{Problem Formulation}
Given a serial nonredundant manipulator, determine if it is cuspidal.

\subsection{Solution Method} \label{sec:ID_soln_method}
If this procedure terminates, the robot is cuspidal:
\begin{enumerate}
    \item Pick a random \new{end effector} pose. \new{This can be done by finding the forward kinematics of a random joint angle vector.}
    \item Find all IK solutions for that pose.
    \item Find all pairs of solutions with the same \(\sign(\det(J))\).
    \item For each pair, find a path by linearly interpolating in joint space and find \(\det(J)\) along the path. If there is a path that never crosses zero then the robot is cuspidal. Else, restart.
\end{enumerate}

Unlike~\cite{marauli2023time}, we only consider \(q_i \in [-\pi, \pi]\) rather than considering both increasing and decreasing joint paths. Unlike~\cite{salunkhe2023trajectory} we did not optimize any singular paths to avoid a singularity.
Despite the simplicity of our algorithm, we were able to quickly prove cuspidality for many robots within a few iterations.

In \cite{salunkhe2024kinematic}, \cut{cuspidality was determined}\new{the authors determined if a robot was cuspidal} by finding all IK solutions using the Husty-Pfurner algorithm\cut{, which took on average 10 ms per end effector pose,} and then connecting them using a multiple shooting approach\cut{, which took 8.52 sec}. \new{Finding all IK solutions took on average 10 ms per end effector pose, and the multiple shooting approach took 8.52 sec.} In contrast, IK for a single end effector pose using IK-Geo takes under 1 ms for robots with at least one pair of intersecting or parallel axes.

\subsection{ABB GoFa CRB 15000 is Cuspidal}

The ABB GoFa~\cite{ABB_gofa} is a new line of cobots with three kinematic variants: 5~kg, 10~kg, and 12~kg. The 5~kg and 1\cut{0}\new{2}~kg variants have intersecting axes~\((1, 2)\). All variants have parallel axes~\((2, 3)\), intersecting axes~\((4, 5)\), and nonconsecutive intersecting axes~\((4, 6)\).
We found examples of non-singular changes of IK solutions for all three variants. To the best of our knowledge, we are the first to report that the GoFa is a cuspidal manipulator.

\begin{figure}[t]
    \centering
    \includegraphics[scale=0.5, clip]{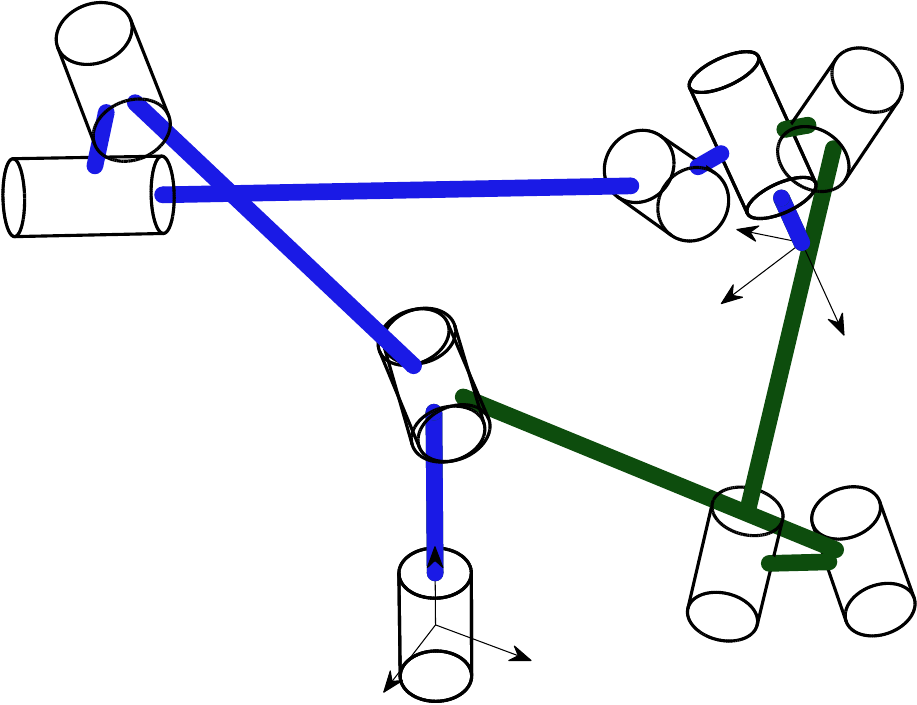}\\[0.1in]
    \includegraphics[scale=0.5, clip]{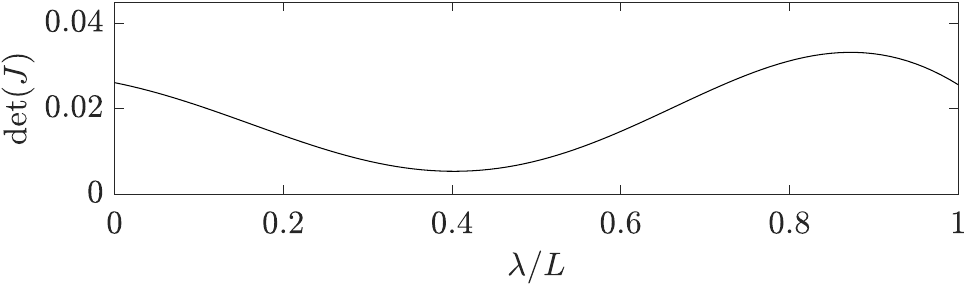}
    \caption{Nonsingular change of solution for the GoFa 5~kg  using a linear joint move from \(q_A\) (left,  blue) to \(q_B\) (right, green).
    }
    \label{fig:gofa_det_J}
\end{figure}

One example nonsingular change of solution for the GoFa 5~kg is shown in Fig.~\ref{fig:gofa_det_J}, which is a \texttt{MoveJ} path between
\begin{equation}
    q_A = \begin{bmatrix}
        -0.8000 \\   \phantom{-}0.5900 \\   \phantom{-}2.3400  \\   \phantom{-}2.7200   \\ \phantom{-}1.0600 \\  -1.8400
    \end{bmatrix},\quad
    q_B = \begin{bmatrix}
        \phantom{-}2.2599  \\  \phantom{-}2.1999 \\   \phantom{-}2.6677  \\  \phantom{-}2.5298  \\ -2.5286  \\  \phantom{-}0.4831
    \end{bmatrix}.
\end{equation}
\new{Although this example exceeds the robot joint limits, we can also find other examples within the joint limits and verify the motion in RobotStudio.}

\subsection{Robots with Three Parallel Axes May Be Cuspidal}
\begin{figure}
    \centering
    \includegraphics[scale=0.5, clip]{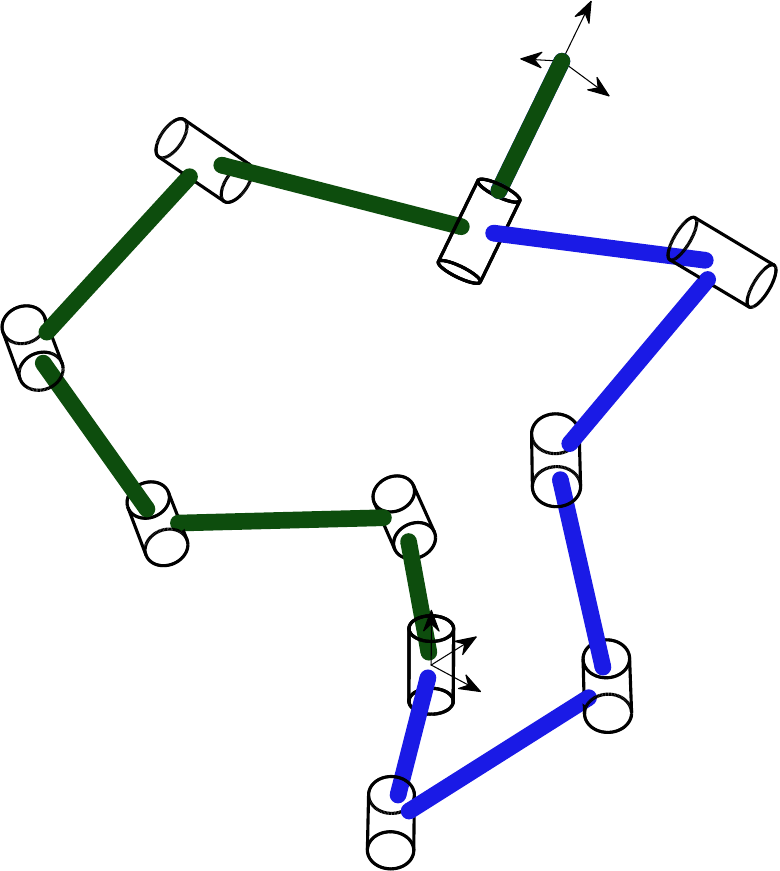} \\[0.1in]
    \includegraphics[scale=0.5, clip]{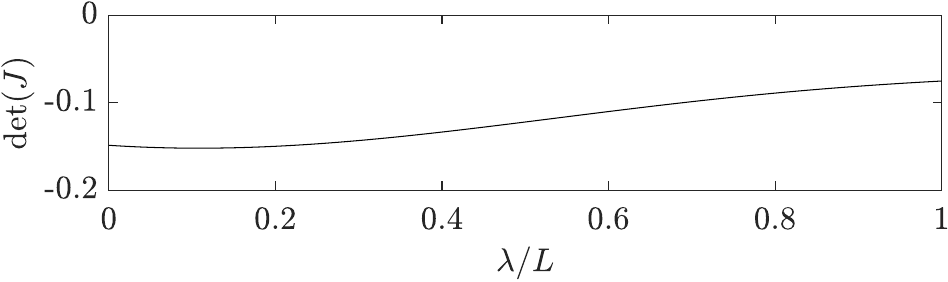}
    \caption{Nonsingular change of solution for a robot with three parallel axes using a linear joint move from \(q_A\) (right,  blue) to \(q_B\) (left, green).}
    \label{fig:3_parallel_bots}
\end{figure}

Others have mentioned that robots with three parallel axes are noncuspidal~\cite{wenger2023review}. Although this is true for common robots such UR cobots which have other intersecting axes and can be solved with a series of quadratics, \cut{in general,} this is false \new{in general}. \cut{It is easy to}\new{One can easily} find a counterexample by searching through random robots and poses and testing if any \texttt{MoveJ} path is nonsingular.
For example, the robot with kinematic parameters
\begin{equation}
\begin{gathered}
    p_{01}=0,\ p_{12} = 0.1e_x + 0.7 e_y,\  p_{23}=p_{34} = p_{45} = 0.7 e_z,\\   p_{56} = 0.3 e_z + 0.9 e_z,\
    p_{6T} = 0.5 e_y\\
    h_1 = ez,\ h_2=h_3=h_4=h_6=e_y, \ h_5 = e_x.
\end{gathered}
\end{equation}
has a nonsingular change of solution (Fig.~\ref{fig:3_parallel_bots}) with a linear joint move between
\begin{equation}
    q_A = \begin{bmatrix}
        -2.4000\\
        -0.9000\\
        \phantom{-}1.1000\\
        -0.8000 \\
        \phantom{-}2.3000 \\
        -1.3000
    \end{bmatrix},\quad
    q_B = \begin{bmatrix}
        \phantom{-}0.9940 \\
        -1.4391\\
        \phantom{-}0.9530\\
        \phantom{-}1.2368\\
        \phantom{-}1.0004\\
        \phantom{-}1.5942
    \end{bmatrix}.
\end{equation}

\section{Path Planning Using Graph Method}\label{sec:path_planning}
Robot operations such as welding may include a prescribed task-space path with the 6-DOF end effector pose fully specified for each point along the path. The path planning problem is to find all possible robot joint-space paths that achieve this task-space path and pick the best one according to some optimization metric.

While the IK problem finds the joint angles for a single end effector pose, the path planning problem extends this to find the entire joint path for an end effector path. \cut{It is not sufficient to apply}\new{Applying} IK to each point in the end effector path \new{is not sufficient} because the joint path must be continuous and may have other constraints, and the IK problem admits multiple solutions.

Many IK solvers including IK-Geo, which we use in our path planning algorithm, and even the Husty-Pfurner method may not perfectly solve the IK problem. For example, IK-Geo may miss an IK solution for robots that are solved with a 1D search if the sampling is not sufficiently dense, and the Husty-Pfurner method may miss a solution if some joint angle is close to \(\pm \pi\) due to the singularity of \(\tan(q_i/2)\). Therefore, the path planning method should be robust to occasional missed IK solutions.

\subsection{Problem Formulation}
Denote an end effector path \(\underline\chi_{0T}(\lambda)\) with path index \(\lambda \in [0, L]\), and denote a point along the path by \(\chi_{0T}(\lambda)\). The goal of the path planning problem is to find the optimal continuous corresponding joint-space path~\(\underline q(\lambda)\).
If desired, the path can be indexed to time with a trajectory~\(\underline\lambda(t)\), where \(t \in [0, T]\).
We can discretize the path into \(\new{K}+1\) evenly spaced points so that \(L = \new{K} \Delta \lambda\). Define the discrete path index \(k \in \{0, 1, \dots, \new{K}\}\) so that \(\chi_{0T}[k]=\chi_{0T}(k\Delta\lambda)\). Similarly, we can represent the joint-space solutions using the discretized path~\(\underline q[k]\).

The optimization metric \(C(\cdot)\) is the sum over the squared finite differences in joint position divided by each time interval:
\begin{align}\label{eq:L2_norm_approx}
    C(\underline q[k]) &=
    \sum_{k=0}^{\new{K}-1} c(q[k+1], q[k]),\\
    c(q_2, q_1) &=     
    \frac{\norm{\wrapToPi(q_2-q_1)}^2}{\Delta \lambda},
\end{align}
where \(\wrapToPi(\Delta q)\) wraps each element of \(\Delta q\) to the range \([-\pi, \pi]\).
In the limit as \(\Delta \lambda \rightarrow 0\) the metric becomes the squared \(L_2\) norm of the joint derivative with respect to~\(\lambda\): 
\begin{equation} \label{eq:vel_norm}
    C(\underline q(\lambda)) = \norm{\underline{q'} (\lambda)}_2^2
    = \int_0^L \norm{\frac{d q(\lambda)}{d \lambda}}^2 d\lambda.
\end{equation}
If \(\lambda\) is the path length measured in m, then 
\(
    (
    C(\underline q(\lambda)) /L
    )^{1/2}\)
is the RMS average of joint movement over path length with units rad/m.

Soft constraints can be included by adding terms to the optimization metric. They may be particularly useful compared to hard constraints when the path planner is used inside an optimization loop as in Section~\ref{sec:path_opt}. Options include:
\begin{itemize}
    \item Joint limits: A barrier function can be included for penalizing \(q\) close to some boundary. If \(\underline \lambda(t)\) is given, then costs can be included for the finite difference approximations of joint velocity or acceleration. (Otherwise, \(\underline\lambda(t)\) can be determined later to ensure any velocity or acceleration constraints are satisfied.)
    \item Singularity avoidance: The kinematic manipulability measure \(\mu = \sqrt{\det(J W J\tr)}\)~\cite{yoshikawa1985manipulability} can be included for each time step, where \(W\) is a joint-space weight matrix.
\end{itemize}
Options for hard constraints on \(\underline q(\lambda)\) include:
\begin{itemize}
    \item Bounds on manipulability, position, or velocity: Similarly to the soft constraint options, bounds may be placed on \(\mu\) or \(q\), and bounds may be placed on joint velocity or acceleration if \(\underline \lambda(t)\) is given.
    \item Nonsingularity: If this constraint is not enforced, a path can be smooth and pass through a singularity because at the singularity large joint motion occurs only if the robot moves in a singular direction.
    \item Repeatability: If the prescribed task-space path is closed, then it may be desirable to constrain the joint-space paths to be repeatable or also regular.

\end{itemize}
Hard constraints are realized by changing the graph construction as detailed in the next section.

\subsection{Graph Method}
For a 6-DOF manipulator, the task-space pose \(\chi_{0T}[k] = (R_{0T}[k], p_{0T}[k])\) consists of the orientation and position of the end effector task frame in the robot base frame.
For each \(\chi_{0T}[k]\) the \cut{inverse kinematics}\new{IK} solutions are \(q^1[k], \ q^2[k], \  \dots, \  q^{M[k]}[k]\), where \(M[k]\) is the number of IK solutions for \(\chi_{0T}[k]\). (A special case is when the IK solutions include a solution at an internal singularity, in which case self-motion is possible at that solution.)
We construct a weighted directed acyclic graph in the following manner:
\begin{itemize}
    \item  Each IK solution \(q^m[k]\) is a represented by a graph vertex \(V_{k,m}\). There are also two special vertices: the start vertex \(S\) and the finish vertex \(F\).
    \item Each IK solution vertex \(V_{k,m}\) has a weighted edge to \(V_{k+1,l}\) if the metric \(c(q^l[k+1], q^m[k]))\)
    is less than some user-defined \(\epsilon = \Delta \lambda \epsilon_0\)\new{.}\cut{, where} \new{If the optimization metric is \eqref{eq:L2_norm_approx} and \(\dot q_{max}\) is the maximum desired joint velocity, then the threshold \(\epsilon_0\) can be set according to} \(\epsilon_0 \approx \norm{\dot q_{max}}^2\).
    \item Vertex \(S\) is connected to each \(V_{0,m}\), and vertex \(F\) is connected to each \(V_{\new{K},m}\) with edges of weight 0 directed from~\(S\) and to~\(F\). 
\end{itemize}

The optimal joint-space path is the shortest graph path between vertices \(S\) and \(F\). The weight of the shortest path is the optimal metric.

\new{Some care must be taken to properly tune the sampling density~\(\Delta \lambda\) and the metric threshold~\(\epsilon_0\). There must be enough samples to reasonably approximate the joint-space paths as piecewise linear approximations, and the threshold must be set to correctly discriminate between connected and disconnected IK solutions.}

\paragraph{Joint Limits} Limits on certain joint angles mean\new{s} that we need to take into account the total number of turns rather than only considering \(q_i \in [-\pi, \pi]\). Care must be taken for cyclical paths so that repeating the path does not cause the joint to exceed the joint limit. Velocity or acceleration constraints can be used to determine if two vertices are connected with an edge (hard constraint) or they can be incorporated into the cost of each edge (soft constraint).

\paragraph{Singular and Nonsingular Paths}
\cutmath{Paths encounter singularities if \(\det(J)=0\). Only nonsingular paths are returned if edges are not made between vertices with different signs for \(\det(J)\).}
\new{A joint-space path encounters a singularity if \(\det(J)=0\) for some point along the path. If edges are not made between vertices with different \(\sign(\det(J))\) then the path planner avoids returning singular paths.
}

\paragraph{Repeatable or Regular Paths}
By finding all connections between nodes at \(k = 0\)  and \(k=\new{K}\) (which represent the same IK solutions), we can make a new directed graph showing connections between IK solutions. Then, we must find graph cycles starting and ending at the same IK solution.

Alternatively, rather than constructing a new graph, we can instead connect each \(V_{0,m}\) to the corresponding \(V_{\new{K},m}\) and find minimum-weight cycles directly.

\paragraph{Robustness to Missed IK Solutions}
To account for missed IK solutions, also connect vertex~\(V_{k,m}\) to vertex~\(V_{k+2,l}\) if the metric 
    \begin{equation}
        \frac{\norm{\wrapToPi(q^l[k+2]-q^m[k])}^2}{2 \Delta \lambda}
    \end{equation}
is less than \(\epsilon\). 
Similarly, \(S\) and \(F\) should be connected to \(V_{1,m}\) and \(V_{\new{K}-1, m}\).
If desired, nodes more than 2 timesteps away may be connected \cut{as desired }similarly. 

Due to the triangle inequality, paths with fewer edges and with larger \(\Delta \lambda\) will have a smaller weight than paths between the same vertices with more edges and with smaller \(\Delta \lambda\). To reduce the error in calculating the total path cost, edges should be added to the graph in multiple passes\cutmath{ starting with the smallest \(\Delta \lambda\), and vertices should be connected with edges with larger \(\Delta \lambda\) only if they vertices are not already connected}. \new{First, only connect vertices with distance \(\Delta \lambda\). Then, connected vertices with distance \(2\Delta \lambda\) only if they are not already connected.}

Another strategy to prevent missed IK solutions when using IK-Geo is to utilize continuous approximate solutions. This allows for Cartesian paths at or even through boundary singularities. This means paths can be at the edge of the workspace or can switch between aspects. Care should be taken to ensure the task-space path deviation is not excessive when using these solutions. Continuous approximate solutions are also useful as part of IK based on 1D search\cut{ to return IK solutions at the edge of the error function branch where the function becomes vertical, in which case the robot may not be singular}. \new{In this case, continuous approximate solutions may occur when the error function has a zero at the edge of its branch, which is where the error becomes vertical. These continuous approximate solutions at the edge of the error function branch are usually not singular IK solutions.}

\subsection{Evaluation}
\begin{figure}[t]
    \centering
    \includegraphics[scale=0.5, clip]{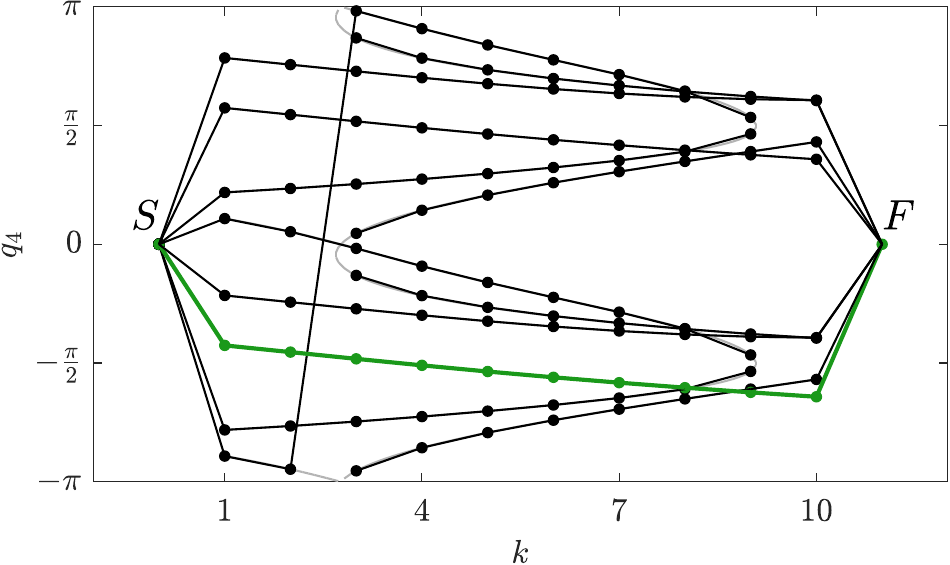}
    \caption{Graph path planning for the FANUC CRX-10iA/L. To find the optimal feasible path, all IK solutions are considered. Vertices representing IK solutions are connected with directed edges weighted by the cost for that movement. Start and finish vertices are also connected to the IK solutions at the beginning and end of the path. The optimal path is found by finding the minimum graph path between the start and finish vertices. An edge crosses the entire graph between \(k=2\) and \(k=3\) because each \(q_i\) is on a torus.\todo{index is supposed to start at 0}}
    \label{fig:graph_demo}
\end{figure}

An example of the graph method for planning a straight-line Cartesian move for the FANUC CRX-10iA/L is shown in Fig.~\ref{fig:graph_demo}. This example uses 10 equally-spaced sample points to make the graph easy to see, but in practice, the density of sample points should be higher. \new{The subtle gray background plot in Fig.~\ref{fig:graph_demo} was created using a higher density of 100 sample points.}

\new{The plots shown in }\cutmath{Furthermore, }Figs.~\ref{fig:cuspidal_topology}, \ref{fig:gofa_MoveL_issues}, and \ref{fig:crx_MoveL_issues} were constructed using the graph method.
\new{All three examples are for straight-line \texttt{MoveL} paths. There are 100 sample points in  Fig.~\ref{fig:3R_infeasible} and 200 sample points in Figs.~\ref{fig:gofa_MoveL_issues} and \ref{fig:crx_MoveL_issues}.

In Section~\ref{sec:path_opt}, we will also demonstrate using the graph method inside an optimization loop. Rather than a simple straight-line path, these examples will use more complicated helical paths each with 500 equally-spaced samples.
}

\section{Path Optimization}\label{sec:path_opt}
\cut{It is common for robots to}\new{Robots commonly} perform operations such as welding where there is freedom in how to fixture the workpiece with respect to the robot, either by moving the workpiece or by moving the robot. In such a scenario, a natural problem to solve is how to find the optimal pose of the workpiece based on some optimization metric.
\subsection{Problem Formulation}

Denote a 6-DOF toolpath defined in the workpiece frame as \(\underline\chi_{PT} [k]\), where \(\chi_{PT} [k] = (R_{PT}[k],\ p_{PT}[k])\), and denote the pose of the workpiece frame in the base frame as \(\chi_{0P} = (R_{0P},p_{0P})\).
The path of the end effector in the base frame is found according to
\begin{equation} \label{eq:toolpath_transformation}
    R_{0T}[k] = R_{0P} R_{PT}[k], \quad
    p_{0T}[k] = p_{0P} + R_{0P} p_{PT} [k].
\end{equation}

Given an optimization metric \(C(\cdot)\), find the optimal workpiece frame pose \(\chi_{0P}\).
This is equivalent to optimizing the pose of the robot base while fixing the toolpath in space.

For 3R robots, \(\underline \chi_{0T} [k]\) includes only the position of the end effector, but the workpiece frame pose \(\chi_{0P}\) still includes both position and orientation.

In Section~\ref{sec:path_planning}, no soft or hard constraints were placed on the end effector path because it was constant. For path optimization, a hard constraint may be placed on \(\chi_{0T}\), or a soft constraint can be added to the optimization metric using a barrier function. If \(\underline \lambda(t)\) is known then constraints can be placed on end effector velocity or acceleration. These task-space constraints can be calculated before finding \(\underline q(\lambda)\).

\subsection{Solution Method}
A locally optimal workpiece frame pose \(\chi_{0P}\) can be found using an optimization method such as \texttt{fminsearch()} in MATLAB~\cite{OptimizationToolbox}, which is a derivative-free local optimizer. To find a feasible starting point, randomly search over \(\mathrm{SO}(3)\) and \(\mathbb{R}^3\) based on the robot workspace until the graph-based path planner finds a feasible \(\underline q[k]\). Then, optimization is performed with the graph-based path planner running at each iteration. If desired, the global minimum can be found with high probability if the optimization is performed with multiple random starting points.

Represent the rotation \(R\new{_{0P}}\) by a unit quaternion with scalar part \(\cos(\theta/2)\) and vector part \(k \sin(\theta/ 2)\), where \(k\) is the axis of rotation and \(\theta\) is the rotation angle.
During optimization, the unit-length constraint does not need to be enforced and the quaternion can be normalized before use\cut{, although}\new{. However,} it should be verified after optimization that the norm does not become too small or large.

Quaternions are used for optimization because they avoid the representation singularities of other parameterizations.
To prevent convergence issues, we require a representation of \(\mathrm{SO}(3)\) where for any point in the representation nearby rotations in \(\mathrm{SO}(3)\) are also nearby in the representation.
In contrast, \(x\)--\(y\)--\(z\) Euler angles have a representation singularity at \(\theta_y = \pm \pi/2\), and the axis times angle representation~\(\theta k\) has a singularity at \(\theta = 2\pi\).

\paragraph*{Eliminating Solution Null Space}
If there are no joint limits or position constraints that depend on the first joint angle~\(q_1\) then rotating the path about the first joint axis \(h_1\) does not change the optimization.

Assume without loss of generality that \(h_1=e_z\) and \(p_{01}\), the offset from the base frame to the first link frame, is only in the \(z\) direction. Write \(R_{0P} = R_z \tilde R\) and \(p_{0P} = R_z \tilde R \tilde p\) so we can rewrite~\eqref{eq:toolpath_transformation} as
\begin{equation}
    R_{0T}[k] = R_z \tilde R R_{PT}[k], \quad
    p_{0T}[k] = R_z  \tilde R ( \tilde{p} + p_{PT} [k]).
\end{equation}
The rotation is applied after the position offset, and the extra rotation \(R_z\) does not affect the optimization and can be set to any angle. Optimizing over \((\tilde R, \tilde p)\) and setting the \(z\) component of the rotation axis of \(\tilde R\) to zero reduces the optimization by one degree of freedom.

\cut{It remains to}\new{For the above analysis to be valid, we must} show that any rotation \(R\) can be represented as a rotation about a vector in the \(xy\) plane followed by a rotation about \(z\), that is \(R = R_z R_{xy}\).
Decompose the rotation into \(z\)--\(x\)--\(z\) Euler angles:
\begin{equation}
    R = R_{z,2} R_x R_{z,1}.
\end{equation}
Using \(R_{z,1} R_{z,1}\tr = I\), rewrite as 
\begin{equation}
     R = R_{z,2} R_{z,1} R_{z,1}\tr R_x R_{z,1}.
\end{equation}
The rotations \( R_{z,2} R_{z,1}\) equal a pure \(z\) rotation \(R_z\).
The remaining rotations \(R_{z,1}\tr R_x R_{z,1}\) are a rotation about \(R_{z,1}\tr e_x\) and therefore a rotation about an axis in the \(xy\) plane, \(R_{xy}\).

\begin{figure}[t]
    \begin{subfigure}[t]{0.5\linewidth}
        \centering{
        \includegraphics[scale = 0.5, clip]{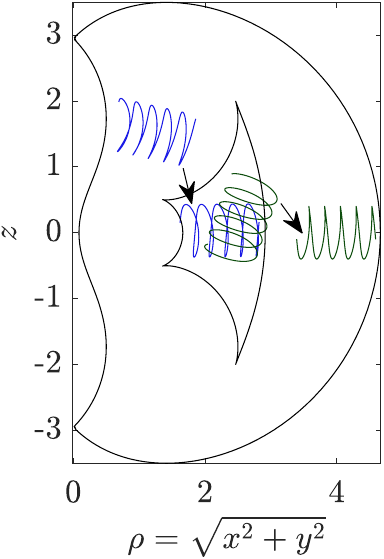}
        \subcaption{\label{fig:3R_opt_rz}}
        }
    \end{subfigure}%
    \begin{subfigure}[t]{0.5\linewidth}
        \centering{
        \includegraphics[scale = 0.5, clip]{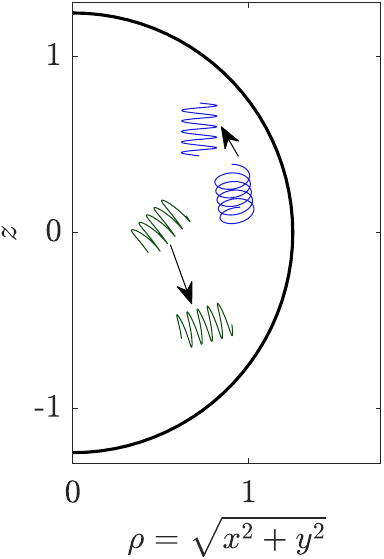}
        \subcaption{\label{fig:crx_opt_rz}}
        }
    \end{subfigure}
    \caption{Initial guesses and locally optimized helix paths. (a)~Cuspidal 3R manipulator. (b)~FANUC CRX-10iA/L. For the CRX, only the workspace envelope is plotted as singularities inside the envelope depend on end effector orientation.
    }
    \label{fig:opt_rz}
\end{figure}

\begin{figure}[t]
    \begin{subfigure}[t]{\linewidth}
        \centering{
        \includegraphics[scale = 0.5, clip]{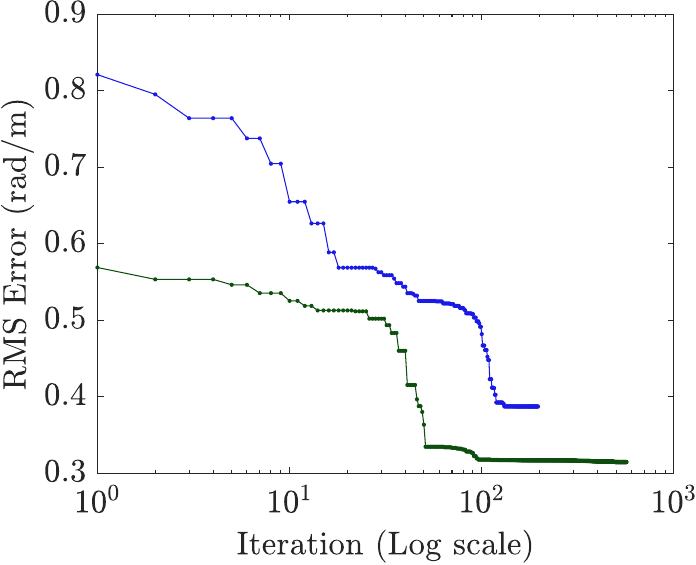}
        \subcaption{\label{fig:opt_history_3R}}
        }
    \end{subfigure}\\[1em]
    \begin{subfigure}[t]{\linewidth}
        \centering{
        \includegraphics[scale = 0.5, clip]{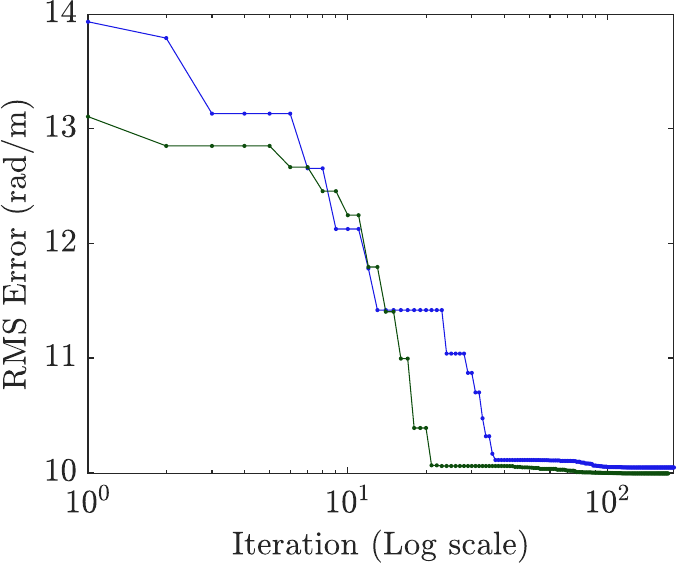}
        \subcaption{\label{fig:opt_history_CRX}}
        }
    \end{subfigure}
    \caption{Optimization error vs iteration. (a)~Cuspidal 3R manipulator. (b)~FANUC CRX-10iA/L.}
    \label{fig:opt_history}
\end{figure}

\begin{figure}[t]
    \begin{subfigure}[t]{\linewidth}
        \centering{
        \includegraphics[scale = 0.5, clip]{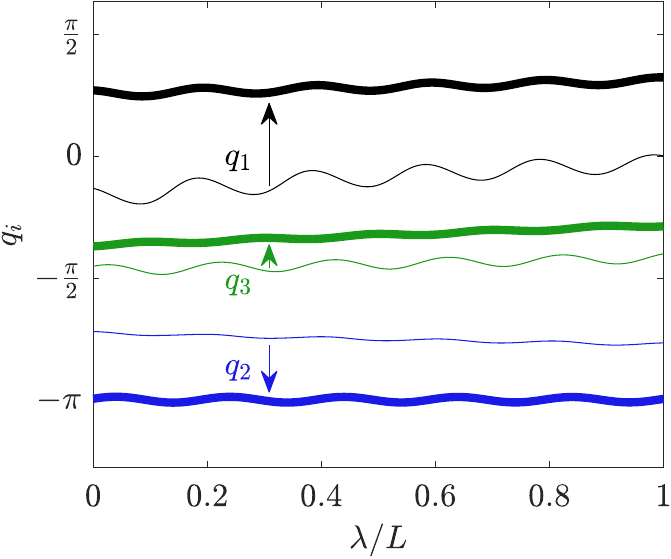}
        \subcaption{\label{fig:opt_before_after_q_3R}}
        }
    \end{subfigure}\\[1em]
    \begin{subfigure}[t]{\linewidth}
        \centering{
        \includegraphics[scale = 0.5, clip]{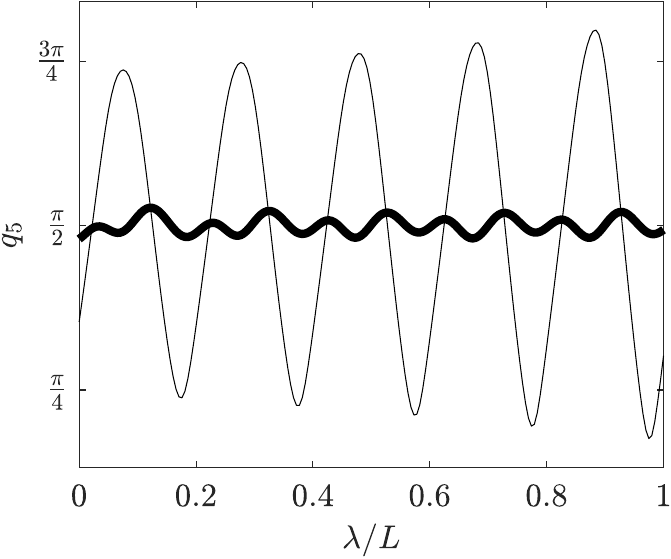}
        \subcaption{\label{fig:opt_before_after_q_CRX}}
        }
    \end{subfigure}
    \caption{Joint-space paths before (thin line) and after (thick line) optimization. (a)~Cuspidal 3R manipulator. (b)~FANUC CRX-10iA/L joint 5.}
    \label{fig:opt_before_after_q}
\end{figure}
\subsection{Evaluation}
Optimization was performed for both a 3R and 6R robot using a helical toolpath \new{consisting of 500 equally-spaced sample points for the path variable \(\lambda\)}. The optimization metric was the \(L_2\) joint velocity norm approximation \new{\eqref{eq:L2_norm_approx}} without any other constraints.

A spatial helical toolpath was generated and optimization was performed for the canonical cuspidal 3R manipulator using two initial guesses.
Results are plotted in Figs.~\ref{fig:3R_opt_rz}, \ref{fig:opt_history_3R}, and \ref{fig:opt_before_after_q_3R}.
The RMS joint velocity average over path length decreased from 0.8209~rad/m to~0.3874 rad/m for one initial guess where the optimized pose was near \(\rho =2\) and from 0.5690~rad/m to 0.3149~rad/m for a different initial guess where the optimized pose was near~\(\rho = 4\).

Optimization was also performed for the FANUC CRX-10iA/L with a 6-DOF helical toolpath and two initial guesses. Results are shown in Figs.~\ref{fig:crx_opt_rz}, \ref{fig:opt_history_CRX}, and \ref{fig:opt_before_after_q_CRX}. RMS joint velocity over path length decreased from 13.9328~rad/m to 10.0447~rad/m for the optimized pose near \(z=0.75\) and decreased from 13.1054~rad/m to 9.9929~rad/m for the optimized pose near \(z=-0.5\).

In both the 3R and 6R cases, there were multiple local minima, demonstrating the importance of using multiple initial guesses.

\section{Conclusion}\label{sec:conclusion}
Most robot practitioners are unaware of cuspidal robots, and many of those who are aware suggest cuspidal robots should be avoided. Our suggestion is that cuspidal robots are difficult to use not because of an inherent problem with the kinematics but because there is a current lack of effective IK and path planning algorithms to control them. Indeed, cuspidal robots offer certain advantages. Wrist offsets may lead to simpler mechanical designs and better dynamical performance.
Such designs even allow for the upper arm to swing past the lower arm, such as for some models in the CRX line. Having multiple IK solutions available in a single aspect can be beneficial because one IK solution may be preferred over another for obstacle or singularity avoidance, and switching between solutions does not require passing through a singularity.
Therefore, it is important not to dismiss cuspidal robots due to their path planning issues but instead to develop new path planning tools so that cuspidal robots can be used to their full potential.

The work presented in this paper is one of the few first steps in improving the path planning capabilities of cuspidal robots. We have shown how to identify cuspidal robots, how to perform path planning given a task-space path, and how to perform path optimization. For the first time, we have shown the ABB GoFa and certain robots with three parallel joint axes are cuspidal. \new{In practice, the identification algorithm can be used to identify the cuspidality of more existing robots and to aid in the design of new robots. }The algorithms presented underscore the importance of IK solvers which can efficiently find all IK solutions, such as IK-Geo. \new{In contrast, we showed RobotStudio does not find all solutions for the ABB GoFa. We demonstrated the path planning and optimization algorithms on both straight-line paths as well as more complicated helix paths. The path planning and optimization algorithms have many practical applications for tasks where the entire toolpath is known and the workpiece or robot base can be moved. This includes subtractive processes such as grinding, sanding, and polishing, and additive processes such as welding, spraying, and 3D printing.}

Future work in improving the path planning and optimization algorithms can include increasing computational efficiency. Path planning efficiency may be increased by using nonuniform sampling along the path to decrease the number of IK solutions evaluated. Path optimization efficiency may be increased by using an analytical gradient of the optimization objective with respect to the toolpath offset.

Future improvements can also focus on new capabilities. The path planning algorithm can include collision avoidance constraints which must consider not only the pose of the end effector but also the pose of each link in the kinematic chain.
The optimization algorithm can be extended to solve different optimization problems such as solving the optimal time problem.

We also raise several general questions about cuspidal manipulators that would be interesting to answer:
\begin{itemize}
    \item What are the cuspidality conditions for robots with three parallel axes? (Conditions have been found for orthogonal robots with three intersecting axes.)
    \item What is a good metric for the ``degree of cuspidality" for a robot? \cut{It is apparent that d}\new{D}ifferent robots have different proportions of random poses with non-singular \texttt{MoveJ} changes of IK solution (I.e., the algorithm in Section~\ref{sec:ID_soln_method} may have a different expected iteration count) and different proportions of random \texttt{MoveL} paths that are infeasible.
    Such a metric may even be useful for noncuspidal robots.
    \new{We hope such a metric can help influence the design or selection of robots to prevent cases of infeasible or nonrepeatible paths.} 
    \item Can we efficiently identify and label the reduced aspect a robot is in? (A reduced aspect is bounded by singularities and characteristic surfaces.) Can we identify how the reduced aspects are connected? This would make path planning even more efficient as we would only have to check the labels along the path, similar to how turn numbers work in configuration parameters for noncuspidal robots.
    \new{\item Can we apply cuspidality analysis to 7-DOF manipulators? Preliminary testing shows that some redundant serial robot arms parameterized by either one joint angle or by the Shoulder-Elbow-Wrist (SEW) angle are cuspidal.}
\end{itemize}
This work in identification, path planning, and path optimization addresses critical theoretical challenges \cut{in}\new{of} cuspidal robots while improving their practical viability. Advancing these methods and addressing the open questions we raise will further unlock the full potential of cuspidal manipulators.
% Check for missing citations
% \nocite{*}

\interlinepenalty=10000 % Don't break in middle of bib entry
\bibliographystyle{asmejour}
\bibliography{bib}
\end{document}